\documentclass{article}

\usepackage{arxiv}

\usepackage[utf8]{inputenc} 
\usepackage[T1]{fontenc}    
\usepackage{hyperref}       
\usepackage{url}            
\usepackage{booktabs}       
\usepackage{amsfonts}       
\usepackage{nicefrac}       
\usepackage{microtype}      
\usepackage{lipsum}
\usepackage{graphicx}
\usepackage{multirow}
\usepackage{amsmath}
\graphicspath{ {./images/} }

\title{DEJIMA: A Novel Large-scale Japanese Dataset for Image Captioning and Visual Question Answering}

\author{
 Toshiki Katsube \\
  The University of Tokyo\\
  \texttt{katsube@mi.t.u-tokyo.ac.jp} \\
  \And
 Taiga Fukuhara \\
  The University of Tokyo\\
  \texttt{fukuhara@mi.t.u-tokyo.ac.jp} \\
  \And
 Kenichiro Ando \\
  RIKEN, The University of Tokyo\\
  \texttt{ando@mi.t.u-tokyo.ac.jp} \\
  \And
 Yusuke Mukuta \\
  The University of Tokyo, RIKEN\\
  \texttt{mukuta@mi.t.u-tokyo.ac.jp} \\
  \And
 Kohei Uehara \\
  The University of Tokyo\\
  \texttt{uehara@mi.t.u-tokyo.ac.jp} \\
  \And
 Tatsuya Harada \\
  The University of Tokyo, RIKEN\\
  \texttt{harada@mi.t.u-tokyo.ac.jp} \\
  \And
}

\begin{document}
\maketitle
\begin{abstract}
  This work addresses the scarcity of high-quality, large-scale resources for Japanese Vision-and-Language (V\&L) modeling. We present a scalable and reproducible pipeline that integrates large-scale web collection with rigorous filtering/deduplication, object-detection-driven evidence extraction, and Large Language Model (LLM)-based refinement under grounding constraints. Using this pipeline, we build two resources: an image–caption dataset (DEJIMA-Cap) and a VQA dataset (DEJIMA-VQA), each containing 3.88M image–text pairs, far exceeding the size of existing Japanese V\&L datasets. Human evaluations demonstrate that DEJIMA achieves substantially higher Japaneseness and linguistic naturalness than datasets constructed via translation or manual annotation, while maintaining factual correctness at a level comparable to human-annotated corpora. Quantitative analyses of image feature distributions further confirm that DEJIMA broadly covers diverse visual domains characteristic of Japan, complementing its linguistic and cultural representativeness. Models trained on DEJIMA exhibit consistent improvements across multiple Japanese multimodal benchmarks, confirming that culturally grounded, large-scale resources play a key role in enhancing model performance. All data sources and modules in our pipeline are licensed for commercial use, and we publicly release the resulting dataset and metadata to encourage further research and industrial applications in Japanese V\&L modeling. The project page is available at \url{https://mil-tokyo.github.io/DEJIMA-dataset}. The dataset with accompanying metadata is released at \url{https://huggingface.co/MIL-UT/DEJIMA-dataset}, and models trained on DEJIMA are publicly available at \url{https://huggingface.co/MIL-UT/DEJIMA-models}.
\end{abstract}


\section{Introduction}
Recent years have seen rapid progress in Vision-and-Language (V\&L) models, which jointly process visual and textual information for tasks such as image captioning, Visual Question Answering (VQA), and image retrieval. The performance of these models critically depends on the availability of high-quality, large-scale datasets. However, most existing V\&L datasets have been constructed primarily in English \cite{nguyen2024multilingual}, and there remains a severe shortage of large, culturally appropriate resources for Japanese language.

Existing Japanese V\&L datasets face clear trade-offs among quality, cultural adequacy, and scalability. Manually annotated datasets such as STAIR Captions \cite{Yoshikawa2017} (164K images) and Japanese Visual Genome \cite{C18-1163} (99K images) provide high-quality but small-scale data. Machine-translated datasets inherit biases and linguistic unnaturalness from English sources, while automatically crawled datasets like Japanese Image Text Pairs \cite{sasagawa2024constructing} achieve large scale but often include noise, poor grounding, or alt-texts that are not natural Japanese sentences. Moreover, existing web-crawled and automatically constructed datasets are limited to caption-style or interleaved image–text formats.

To address these limitations, we propose DEtection-based Japanese Image-text dataset for Multi-modal Analysis (DEJIMA), a scalable pipeline for constructing culturally grounded, high-quality Japanese V\&L datasets. Our approach integrates:
(1) \textbf{web collection, filtering, and deduplication} of images and corresponding alt-texts from Japanese web pages in Common Crawl\footnote{\url{https://commoncrawl.org/}};
(2) \textbf{object detection} for extracting visually verifiable evidence; and
(3) \textbf{LLM-based refinement} that enforces fluency and factual grounding under explicit constraints.

Using this pipeline, we created two resources: DEJIMA-Cap (image–caption pairs) and DEJIMA-VQA (image–question–answer triples), each containing approximately 3.88 million entries. This corresponds to 23.7× the scale of STAIR Captions and 39.1× that of Japanese Visual Genome. All models used in our pipeline are under commercially permissible licenses, enabling the resulting dataset to be used safely for research and commercial purposes.

Human evaluations demonstrate that DEJIMA achieves markedly higher scores in Japaneseness and linguistic naturalness compared with both manual and translation-based baselines, while maintaining factual correctness at a level comparable to human-annotated data. These results indicate that DEJIMA successfully balances scale, cultural richness, and quality. Furthermore, V\&L models trained on DEJIMA outperform prior Japanese datasets on benchmarks such as JA-VLM-Bench-In-the-Wild \cite{akiba2024evomodelmerge} and heron-bench \cite{inoue2024heron}, highlighting the effectiveness of our detection-guided and LLM-refined approach.

In summary, our contributions are threefold:
\begin{itemize}
  \item We introduce a detection-based Large Language Model (LLM) refinement pipeline that enhances fluency and factual grounding through evidence-first generation.
  \item We construct a large-scale Japanese Vision-and-Language (V\&L) dataset (3.88M pairs) that surpasses existing resources by over 20× in scale while preserving cultural and linguistic naturalness.
  \item We demonstrate that models trained with DEJIMA achieve strong performance and natural, culturally coherent Japanese text generation, approaching human annotation quality.
\end{itemize}

A V\&L dataset comprises image–text pairs and is used to train V\&L models. Four common construction paradigms are: (1) human annotation, (2) web crawling, (3) translation from other languages, and (4) generation/augmentation with modern generative models.

\paragraph{Human-annotated Datasets.}
Manual annotation is costly but offers strong image–text alignment and high textual quality. English resources include Microsoft COCO Captions \cite{lin2014microsoft}, Flickr30k \cite{young2014image}, Visual Genome \cite{krishna2017visual}, and GQA \cite{hudson2019gqa}. Notably, building MS COCO required over 70{,}000 crowd-worker hours. Japanese counterparts include STAIR Captions \cite{Yoshikawa2017} and Japanese Visual Genome \cite{C18-1163}.

\paragraph{Web-crawled Datasets.}
Large-scale efforts collect images and their associated alt-text from the web: CC3M \cite{sharma2018conceptual}, CC12M \cite{changpinyo2021conceptual}, and LAION-400M \cite{schuhmann2021laion}. For Japanese, the Japanese Image Text Pairs dataset \cite{sasagawa2024constructing} is notable. Despite the large scale, alt-text can be noisy (partial descriptions, mismatched semantics, or unnatural sentences).

\paragraph{Translated Datasets.}
English resources are often translated to other languages (e.g., LLaVA-Instruct-150K-JA\footnote{\url{https://huggingface.co/datasets/turing-motors/LLaVA-Instruct-150K-JA}} from LLaVA Visual Instruct 150K \cite{liu2023visual}). While scalable, translations can introduce unnatural phrasing and attenuate culture-specific knowledge.

\paragraph{Generated/Augmented Datasets.}
LLM-based augmentation can rewrite captions to enrich linguistic variety (e.g., LaCLIP \cite{fan2023improving}), whereas fully synthetic pipelines (e.g., SynthVLM \cite{liu2024synthvlm}) can generate image–text pairs from scratch. Risks include hallucinations when text is detached from image evidence and cultural bias if generators were not trained on diverse, multilingual sources.

\section{Dataset}
\begin{figure}[t]
  \begin{center}
    \includegraphics[width=0.5\textwidth]{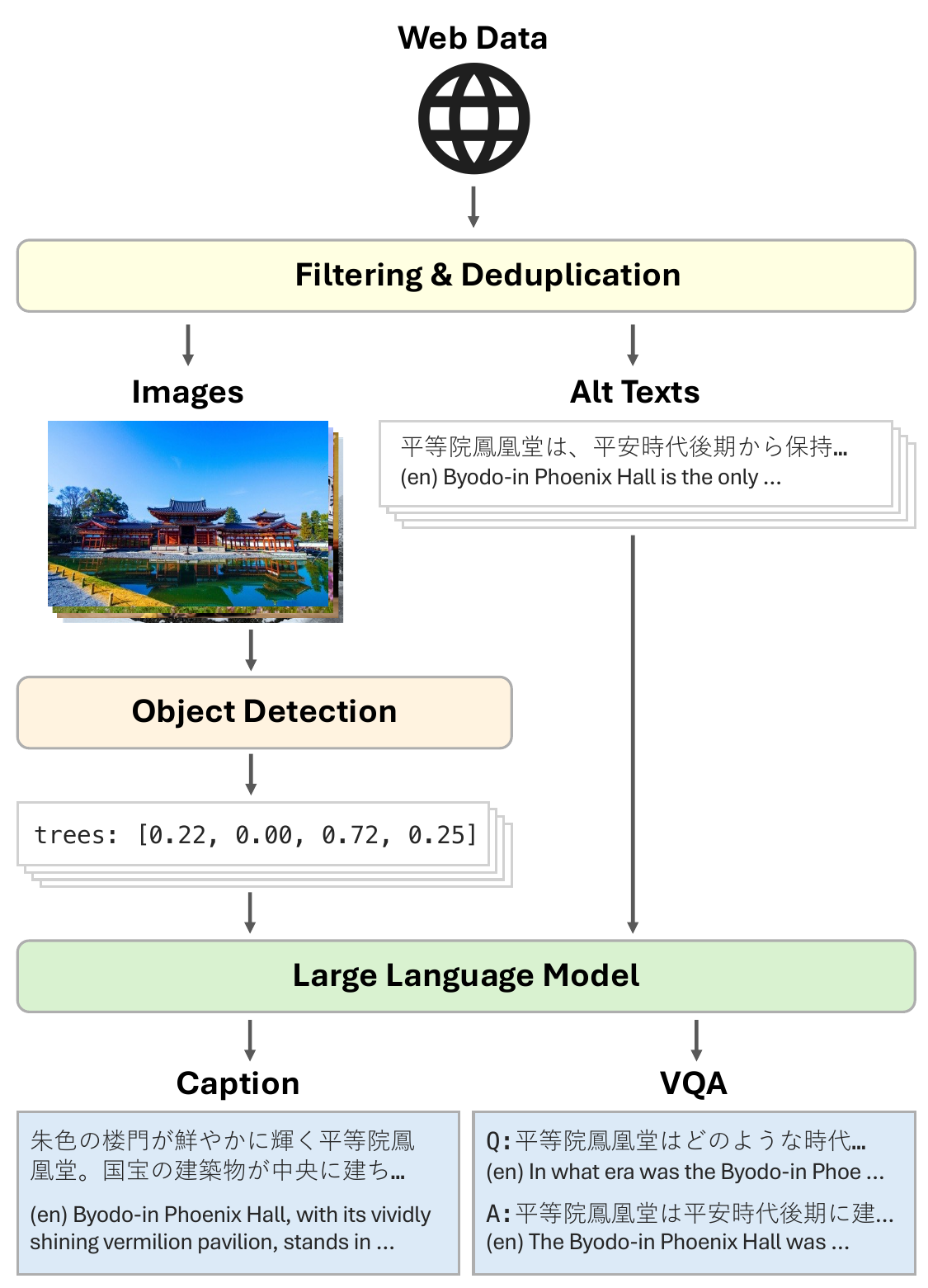}
    \caption{Pipeline: (1) web collecting with strict filtering \& deduplication, (2) detection-driven evidence extraction, (3) LLM refinement with grounding/safety constraints.}
    \label{pipeline}
  \end{center}
\end{figure}

Figure~\ref{pipeline} outlines our three-stage pipeline: (1) collecting URLs and alt-text from the web with rigorous filtering and deduplication, (2) object detection and tag calibration, and (3) LLM-based caption/VQA generation with grounding constraints.

\subsection{Creation Pipeline}
\paragraph{Stage 1: Web Collection, Filtering, and Deduplication.}
We use Common Crawl (pages crawled Aug 7, 2022–Jan 26, 2025; \(\sim\)50B pages) as the source and retain only Japanese pages and safe content through a multi-step filtering pipeline.
Language and adult filtering are conducted using \texttt{HojiChar}\footnote{\url{https://github.com/HojiChar/HojiChar}}, applying \texttt{AcceptJapanese()} to keep only Japanese text and \texttt{DiscardAdultContentJa()} to exclude adult material.
Text normalization is performed (e.g., converting full-width spaces to half-width, trimming, and compressing consecutive spaces), and only entries whose alt-text passes all filters are kept together with their image URLs.

Highly frequent alt-texts are then removed to promote diversity by discarding entries with frequency \(\geq 10\).
For near-duplicate removal, we compute perceptual hashes (\texttt{imagehash.phash}) and deduplicate by the pair \((\texttt{pHash}, \texttt{alt})\), retaining only the first occurrence to eliminate visually identical images with the same caption.

To further ensure semantic alignment, we calculate the cosine similarity between image and alt-text embeddings using the Japanese CLIP model \texttt{line-corporation/clip-japanese-base}\footnote{\url{https://huggingface.co/line-corporation/clip-japanese-base}}.
Only entries above the 30th percentile of the global similarity distribution are kept, removing the noisiest bottom \(\sim30\%\).
Each retained pair is stored with its alignment score.
Finally, we exclude images with unsupported extensions (\texttt{jpeg}, \texttt{jpg}, \texttt{png} only), filter out extreme aspect ratios (>$3{:}1$), and download the remaining images via \texttt{img2dataset}\footnote{\url{https://github.com/rom1504/img2dataset}}.

\paragraph{Stage 2: Object Detection and Tag Calibration.}
Visual evidence is extracted using Recognize Anything (RAM) \cite{zhang2024recognize} with the RAM++(14M) checkpoint\footnote{\url{https://huggingface.co/xinyu1205/recognize-anything-plus-model/blob/main/ram_plus_swin_large_14m.pth}}, which provides comprehensive object tags.
Bounding boxes for each detected object are then obtained using Grounding DINO \cite{liu2024grounding}, ensuring spatial grounding of visual entities.
This combination yields both rich object-level coverage and accurate localization information for downstream text generation.

\paragraph{Stage 3: LLM-guided Generation with Grounding Constraints.}
We refine or generate texts using \texttt{qwen2.5-bakeneko-32b}\footnote{\url{https://huggingface.co/rinna/qwen2.5-bakeneko-32b}}, conditioning on both alt-text and detection outputs (labels and bounding boxes) to provide explicit visual grounding.
A few-shot prompting setup is employed, in which curated exemplars guide the model toward culturally natural and semantically faithful text generation.

For the caption dataset, the LLM refines noisy alt-texts into fluent and descriptive Japanese captions consistent with detected objects.
For the VQA dataset, both questions and answers are generated by the LLM under the same grounding signals, ensuring coherence between the visual evidence and linguistic reasoning.

\paragraph{Efficiency Comparison.}
While manually annotated datasets such as MS~COCO (123K images) required over 70{,}000 crowd-worker hours to complete \cite{lin2014microsoft}, our automated DEJIMA pipeline generates approximately 1~million captions or VQA pairs in about 40~GPU hours using NVIDIA~A100 GPU. This represents an efficiency gain of several orders of magnitude.

\paragraph{Dataset Variants.}
We prepare multiple dataset variants to isolate the contributions of each component in our pipeline.
For the captioning task, we construct four datasets:
\textbf{DEJIMA-Cap-Simple} consists of raw image–alt-text pairs after filtering;
\textbf{DEJIMA-Cap-Refined} uses only LLM-refined alt-texts;
\textbf{DEJIMA-Cap-Detection} contains captions generated by the LLM from only detection tags;
and \textbf{DEJIMA-Cap-All} integrates both alt-texts and detection tags as inputs to the LLM.

Similarly, for the VQA task, we prepare three corresponding datasets:
\textbf{DEJIMA-VQA-Refined}, \textbf{DEJIMA-VQA-Detection}, and \textbf{DEJIMA-VQA-All}.

\begin{figure}[t]
  \centering
  \includegraphics[width=0.5\textwidth]{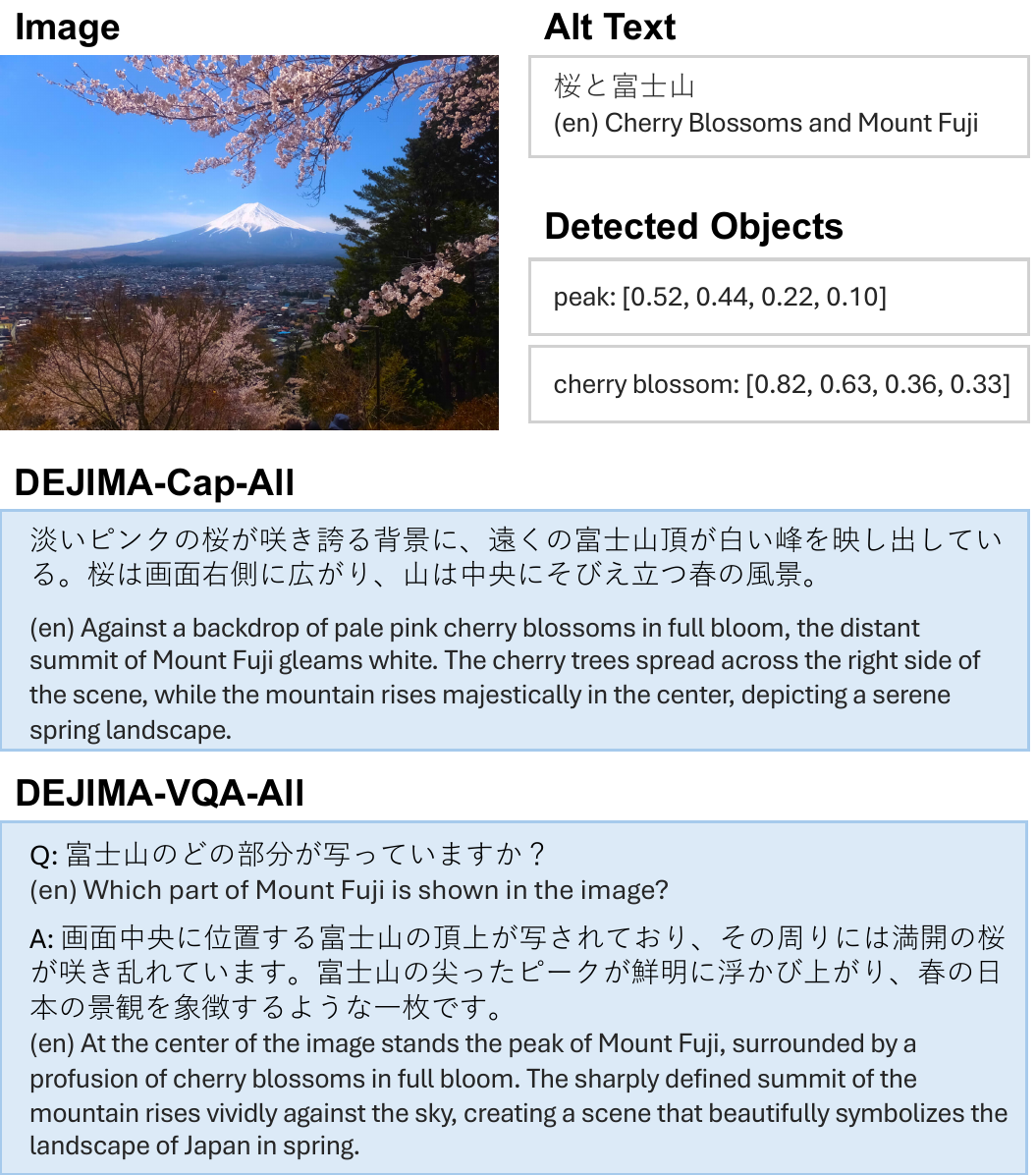}
  \caption{
    Example instances from DEJIMA-Cap and DEJIMA-VQA.
    The figure illustrates the pipeline flow:
    (1) the input image is processed by the object-detection model to obtain \emph{detected objects} (labels and bounding boxes);
    (2) the \emph{alt text} and detected objects are provided as inputs to the LLM, which generates a refined \emph{caption} and grounded \emph{VQA} pairs in Japanese. English translations are also included in the figure.
  }
  \label{fig:dataset_example_card}
\end{figure}

\paragraph{Dataset Example.} Figure~\ref{fig:dataset_example_card} shows one example each from DEJIMA-Cap and DEJIMA-VQA.

\subsection{Dataset Analysis}
\label{sec:dataset-analysis}
\paragraph{Statistical Comparison.}
Table~\ref{tab:stat_comparison} provides an overview of the quantitative properties of existing Japanese Vision–Language datasets and our DEJIMA resources. For reference, we also include MS~COCO~Translation and GQA~Translation, which are machine-translated versions of the original English datasets (MS~COCO~Captions and GQA) produced using \texttt{qwen2.5-bakeneko-32b}.

From a scalability perspective, DEJIMA-Cap and DEJIMA-VQA each comprise roughly 3.88~million image–text pairs—23.7~times larger than STAIR~Captions (164K images) and 39.1~times larger than Japanese~Visual~Genome (99K images). This vast scale allows DEJIMA to cover a substantially broader range of visual scenes and linguistic phenomena than prior Japanese datasets.

In terms of linguistic diversity, captions in DEJIMA-Cap-All are longer and more informative, with an average length of 79.6~characters, compared to around 21~characters in STAIR~Captions and MS~COCO~Translation. The vocabulary also expands significantly, reaching approximately 287K unique types, indicating rich lexical variety and descriptive expressiveness.

From DEJIMA-Cap-Simple to DEJIMA-Cap-Refined, the average caption length nearly doubles (18 → 38 characters), showing that LLM refinement enhances fluency and expressiveness.
Incorporating object-detection information (DEJIMA-Cap-Detection) further adds visually grounded terms, and combining both detection results and alt-text information (DEJIMA-Cap-All) leads to larger vocabulary expansion (from about 30K to 280K types).
Interestingly, the vocabulary size of DEJIMA-Cap-All becomes slightly smaller than that of DEJIMA-Cap-Simple, which can be attributed to the LLM refinement process filtering out overly specific or uncommon proper nouns that appeared in raw web-derived alt-texts.
This indicates that the inclusion of web-derived alt-text contributes a rich variety of linguistic expressions—such as specific object names, proper nouns, and contextual descriptions—while the LLM-based refinement selectively retains more general and natural ones.
A similar pattern is observed for the VQA data.
These statistics collectively demonstrate that our pipeline not only scales up dataset size but also improves linguistic diversity and descriptive depth by effectively combining visual grounding and naturally written web text.

\begin{table}[t]
  \centering
  \resizebox{\linewidth}{!}{
    \begin{tabular}{lrrrrr}
      \toprule
      Dataset                     & Type                  & \# Images     & \# Texts      & Avg. \# Chars & Vocabulary Size \\
      \midrule
      \multicolumn{6}{l}{\textit{\textbf{Caption}}}                                                                         \\
      STAIR Captions              & Human-annotated       & 123{,}287     & 616{,}435     & 23.80         & 30{,}195        \\
      MS COCO Translation         & Machine-translated    & 123{,}287     & 616{,}767     & 22.41         & 32{,}960        \\
      DEJIMA-Cap-Simple (Ours)    & Alt                   & 3{,}884{,}632 & 3{,}884{,}632 & 18.21         & 336{,}924       \\
      DEJIMA-Cap-Refined (Ours)   & Alt + LLM             & 3{,}884{,}629 & 3{,}884{,}629 & 38.03         & 314{,}900       \\
      DEJIMA-Cap-Detection (Ours) & Detection + LLM       & 3{,}884{,}632 & 3{,}884{,}632 & 49.55         & 30{,}674        \\
      DEJIMA-Cap-All (Ours)       & Alt + Detection + LLM & 3{,}884{,}632 & 3{,}884{,}632 & 79.62         & 287{,}434       \\
      \midrule
      \multicolumn{6}{l}{\textit{\textbf{VQA}}}                                                                             \\
      Japanese Visual Genome      & Human-annotated       & 99{,}208      & 793{,}664     & 19.50         & 20{,}797        \\
      GQA Translation             & Machine-translated    & 71{,}067      & 3{,}999{,}765 & 22.58         & 11{,}856        \\
      DEJIMA-VQA-Refined (Ours)   & Alt + LLM             & 3{,}875{,}343 & 3{,}875{,}343 & 56.62         & 321{,}720       \\
      DEJIMA-VQA-Detection (Ours) & Detection + LLM       & 3{,}883{,}943 & 3{,}883{,}943 & 77.00         & 31{,}929        \\
      DEJIMA-VQA-All (Ours)       & Alt + Detection + LLM & 3{,}882{,}892 & 3{,}882{,}892 & 108.86        & 278{,}860       \\
      \bottomrule
    \end{tabular}
  }
  \caption{Statistical comparison of Japanese V\&L Datasets.}
  \label{tab:stat_comparison}
\end{table}

\paragraph{Representational Coverage.}
To examine the representational coverage of DEJIMA relative to existing datasets, we analyzed the 2D feature distributions obtained by applying PCA to CLIP \cite{ilharco_gabriel_2021_5143773} image embeddings.
All datasets were jointly projected into a shared two-dimensional space, and a common $60\times60$ grid (with 2\% padding) was used to discretize the plane into probability maps $p_d(i,j)$ for each dataset~$d$.

For each dataset, we computed two complementary measures:
(1) the asymmetric \textit{coverage rate} $\mathrm{Coverage}(P|Q)=\sum_{b\in\mathrm{occ}_Q}p_P(b)$,
which quantifies how much probability mass of dataset~$P$ lies within the bins occupied by~$Q$,
and (2) the bidirectional Kullback–Leibler divergences $\mathrm{KL}(P||Q)$ and $\mathrm{KL}(Q||P)$,
computed with numerical stabilization ($\varepsilon=10^{-12}$).
These metrics respectively capture the spatial overlap and the distributional divergence between datasets in the shared embedding space.

Using the domestic dataset \textit{recruit-jp}\footnote{\url{https://huggingface.co/datasets/recruit-jp/japanese-image-classification-evaluation-dataset}} as the reference target,
DEJIMA achieved the highest coverage rate $\mathrm{Coverage}(\text{target}|\text{DEJIMA})=0.785$,
substantially exceeding Japanese Visual Genome (0.435), STAIR Captions (0.430), MS~COCO (0.406), and GQA (0.342).
This indicates that DEJIMA spans approximately 79\% of the visual domain occupied by real Japanese imagery.
Conversely, when measuring $\mathrm{Coverage}(\text{dataset}|\text{target})$,
Japanese Visual Genome exhibited the highest value (0.534), followed by MS~COCO (0.502) and STAIR Captions (0.492),
while DEJIMA scored lower (0.192), suggesting broader support beyond the domestic domain.

The KL divergences show a consistent pattern:
$\mathrm{KL}(\text{recruit-jp}||\text{DEJIMA})=6.03$ (lowest among all pairs),
followed by Japanese Visual Genome ($\approx$12.2), STAIR ($\approx$12.3), MS~COCO ($\approx$12.8), and GQA ($\approx$14.2).
In the reverse direction, $\mathrm{KL}(\text{DEJIMA}||\text{recruit-jp})=16.4$,
indicating that DEJIMA includes additional regions not present in recruit-jp.

\begin{figure}[t]
  \centering
  \includegraphics[width=0.5\textwidth]{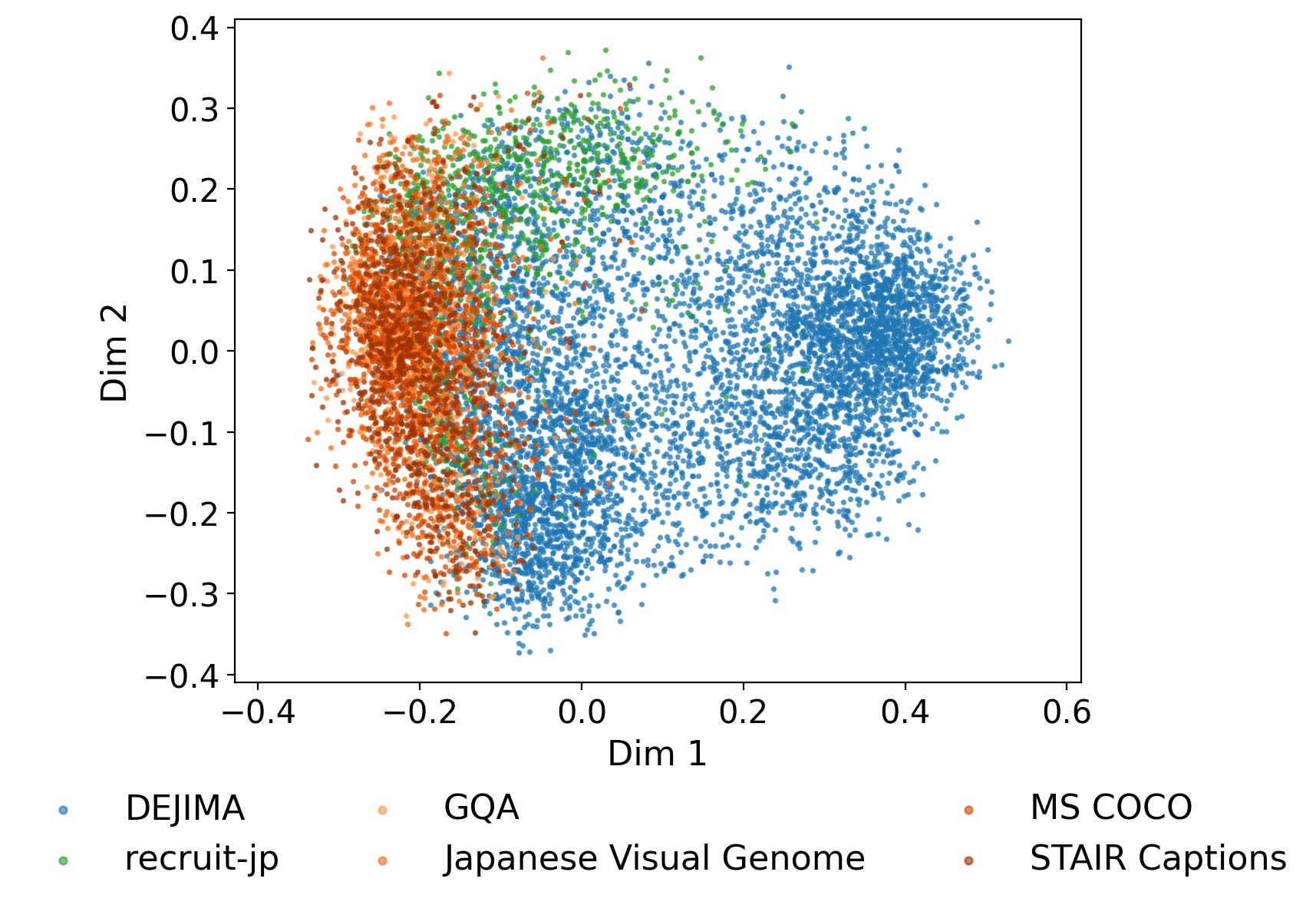}
  \caption{PCA projection of CLIP image embeddings.
    DEJIMA covers the entire region of recruit-jp while extending into broader global contexts.}
  \label{fig:image_pca}
\end{figure}

Figure~\ref{fig:image_pca} visualizes these relationships.
DEJIMA exhibits the widest and most continuous distribution among the compared datasets,
covering the full domestic (Japanese) domain while extending smoothly toward broader, globally oriented image clusters.
These results collectively demonstrate that DEJIMA provides both dense coverage of Japanese visual concepts and expanded representational diversity beyond existing Japanese datasets.

\paragraph{Human Evaluation.} We also conducted human evaluations to assess the quality of the caption and VQA datasets.
We conducted pairwise human evaluations on 150 random samples per dataset. Annotations were collected via crowdsourcing (80 workers total), comparing two datasets side-by-side with randomized order. To ensure quality, we inserted a control item for each worker and excluded inconsistent annotations (22.5\% of workers); final sample sizes \(n\) are reported in each table.

Before the evaluation, all workers were presented with detailed descriptions of the evaluation metrics.
For caption evaluation, the metrics included \emph{Japaneseness of image}, which measures how strongly the image reflects Japanese culture or scenery; \emph{Japaneseness of text}, which captures the presence of Japan-specific expressions or contexts in the caption; \emph{Naturalness of text}, which assesses the fluency and naturalness of the Japanese language; \emph{Image–text consistency}, which evaluates factual alignment between the caption and visible evidence such as color, number, or position; \emph{Coverage}, which measures how well the caption mentions major objects or regions in the image; and \emph{Expressiveness}, which evaluates the richness and vividness of description beyond simple enumeration.
For VQA evaluation, the metrics included \emph{Japaneseness of image}; \emph{Japaneseness of text} and \emph{Naturalness of text} for both question and answer; \emph{Q–A relevance}, which measures whether the answer appropriately addresses the question’s intent; \emph{Q–image consistency}, which assesses whether the question is grounded in the image; and \emph{Answer correctness}, which checks whether the answer is factually correct given the image content.

Two-sided binomial tests were conducted to determine whether observed preference rates significantly deviated from chance (50\%).
Statistical significance was evaluated at the 5\% level, and the Holm–Bonferroni correction was applied across metrics within each comparison.
In the following tables, * indicates significance at the 5\% level.

Tables~\ref{tab:Caption-DEJIMA-Cap-combined} and \ref{tab:VQA-DEJIMA-VQA-combined} show the results of pairwise human evaluations comparing our datasets with existing ones.
Across most metrics, DEJIMA-Cap-All and DEJIMA-VQA-All were consistently preferred, with statistically significant differences in nearly all linguistic and cultural dimensions.

For captions, compared with existing datasets such as MS COCO Translation and STAIR Captions, DEJIMA-Cap-All achieved clearly higher preference rates for \emph{Japaneseness of image} and \emph{Japaneseness of text}.
This improvement stems from the fact that existing datasets are either translations of English captions or annotations on English-centric image resources, while DEJIMA collects images from Japanese websites, naturally reflecting local culture and scenery.
DEJIMA also outperformed these baselines in \emph{Expressiveness}, showing that combining alt-text and detection enables more multi-perspective and vivid descriptions.
However, \emph{image–text consistency} was lower than in human-annotated datasets, reflecting the continued advantage of manual alignment in grounding accuracy.
In \emph{Naturalness of text}, DEJIMA surpassed MS COCO Translation, likely because Japanese LLM generation yields more fluent expressions than machine translation or crowdsourced captions.
Ablation comparisons confirm that excluding either alt-text or detection reduces overall quality—the two inputs complement each other, balancing contextual richness and grounding precision.

For VQA, DEJIMA-VQA-All also outperformed translation-based datasets (GQA Translation) and human-annotated ones (Japanese Visual Genome) in \emph{Japaneseness of image}, \emph{Japaneseness of text}, and \emph{Naturalness of text}.
Again, this advantage arises from DEJIMA’s use of Japanese web imagery, which better captures domestic cultural elements.
Compared with Japanese Visual Genome, DEJIMA-VQA-All performed lower in \emph{Q–A relevance} and \emph{Answer correctness}, showing that human annotation remains more precise for factual grounding.
In contrast, against GQA Translation, DEJIMA showed no significant difference in \emph{Q–A relevance} or \emph{Answer correctness}, and no gap in \emph{Q–image consistency}, demonstrating that high-quality Japanese VQA data can be constructed automatically when grounded by detection.
Ablation results again show that alt-text and detection each compensate for the other’s weaknesses.

\begin{table}[t]
  \centering
  \resizebox{\linewidth}{!}{
    \begin{tabular}{llllllll}
      \toprule
      Compared Dataset     & $n$ & \shortstack{Japaneseness                                                                                                                                        \\of image}    & \shortstack{Japaneseness\\of text}     & \shortstack{Naturalness\\of text}      & \shortstack{Image-text\\consistency}   & Coverage                 & Expressiveness           \\
      \midrule
      MS COCO Translation  & 105 & 82.86\textsuperscript{*} & 87.62\textsuperscript{*} & 86.67\textsuperscript{*} & 20.00\textsuperscript{*} & 39.05\textsuperscript{*} & 92.38\textsuperscript{*} \\
      STAIR Captions       & 135 & 74.07\textsuperscript{*} & 77.78\textsuperscript{*} & 62.22\textsuperscript{*} & 20.74\textsuperscript{*} & 43.70                    & 68.89\textsuperscript{*} \\
      DEJIMA-Cap-Refined   & 105 & --                       & 86.67\textsuperscript{*} & 64.76\textsuperscript{*} & 52.38                    & 61.90\textsuperscript{*} & 91.43\textsuperscript{*} \\
      DEJIMA-Cap-Detection & 135 & --                       & 76.30\textsuperscript{*} & 65.93\textsuperscript{*} & 62.96\textsuperscript{*} & 70.37\textsuperscript{*} & 81.48\textsuperscript{*} \\
      \bottomrule
    \end{tabular}
  }
  \caption{Caption: Pairwise preference of \textbf{DEJIMA-Cap-All} vs. baselines. * indicates significance at 5\%.}
  \label{tab:Caption-DEJIMA-Cap-combined}
\end{table}

\begin{table}[t]
  \centering
  \resizebox{\linewidth}{!}{
    \begin{tabular}{llllllll}
      \toprule
      Compared Dataset       & $n$ & \shortstack{Japaneseness                                                                                                                                        \\of image}    & \shortstack{Japaneseness\\of text}     & \shortstack{Naturalness\\of text}      & \shortstack{Q-A\\relevance}            & \shortstack{Q-Image\\consistency}      & \shortstack{Answer\\correctness}       \\
      \midrule
      GQA Translation        & 105 & 91.43\textsuperscript{*} & 92.38\textsuperscript{*} & 89.52\textsuperscript{*} & 41.90                    & 51.43                    & 41.90                    \\
      Japanese Visual Genome & 135 & 92.59\textsuperscript{*} & 87.41\textsuperscript{*} & 78.52\textsuperscript{*} & 31.85\textsuperscript{*} & 38.52\textsuperscript{*} & 34.81\textsuperscript{*} \\
      DEJIMA-VQA-Refined     & 90  & --                       & 76.67\textsuperscript{*} & 74.44\textsuperscript{*} & 57.78                    & 56.67                    & 57.78                    \\
      DEJIMA-VQA-Detection   & 120 & --                       & 79.17\textsuperscript{*} & 72.50\textsuperscript{*} & 61.67\textsuperscript{*} & 65.83\textsuperscript{*} & 63.33\textsuperscript{*} \\
      \bottomrule
    \end{tabular}
  }
  \caption{VQA: Pairwise preference of \textbf{DEJIMA-VQA-All} vs. baselines. * indicates significance at 5\%.}
  \label{tab:VQA-DEJIMA-VQA-combined}
\end{table}

\section{Experiments}
To examine how much the constructed dataset contributes to improving VLM performance, we trained Vision-and-Language Models (VLMs) using our dataset.

\subsection{Setup}
Following the two-stage training scheme of LLaVA \cite{liu2023visual}, we use a \texttt{siglip-so400m-patch14-384}\footnote{\url{https://huggingface.co/google/siglip-so400m-patch14-384}} vision encoder and \texttt{llm-jp/llm-jp-3-7.2b}\footnote{\url{https://huggingface.co/llm-jp/llm-jp-3-7.2b}} as the language model.
In the first stage, we train the model to align images and captions using the caption datasets, and in the second stage, we fine-tune it on VQA data to enhance multimodal reasoning and grounding abilities.
We evaluate the resulting models on two representative Japanese multimodal benchmarks: JA-VLM-Bench-In-the-Wild \cite{akiba2024evomodelmerge} and Heron-bench \cite{inoue2024heron}.

\paragraph{Benchmarks.}
\textbf{JA-VLM-Bench-In-the-Wild} is a benchmark developed by Sakana AI to evaluate Japanese VLMs in culturally grounded, open-domain scenarios.
It measures the model’s ability to understand and describe Japanese cultural and social contexts through diverse real-world images.
The dataset is constructed through a semi-automatic process:
GPT-4V generates questions and answers for 42 curated Japanese images,
and human annotators filter and refine them.
Thus, it primarily tests linguistic fluency, cultural knowledge, and contextual relevance in free-form Japanese.

\textbf{Heron-bench} was created by Turing Corporation and focuses on more fine-grained multimodal reasoning.
It contains 21 carefully selected Japanese images across seven domains (e.g., anime, art, food, culture, landscape, landmark, transportation),
with 102 manually written questions by researchers.
It emphasizes precise perception, factual correctness, and reasoning grounded in Japanese context,
making it a stricter test for factual alignment and multimodal understanding.

\subsection{Benchmark Results}
Table~\ref{tab:experimental_results_actual} reports model performance on both benchmarks (LLM-as-a-Judge scores; higher is better).

The divergent results on the two benchmarks highlight a critical relationship between the granularity of the training data and the specific demands of the evaluation task.

On JA-VLM-Bench-In-the-Wild, the top performance of the DEJIMA-Cap-Simple \& DEJIMA-VQA-Refined model is instructive. This benchmark primarily assesses a model's ability to handle general-purpose queries about widely recognizable Japanese subjects, such as traditional clothing (e.g., kimono) or major landmarks (e.g., Tokyo Tower). For this purpose, the concise, natural language of raw alt-texts in DEJIMA-Cap-Simple appears sufficient to build a strong foundational image-text alignment. The highly detailed and lengthy captions of DEJIMA-Cap-All, while factually rich, may have inadvertently biased the model toward object-level specifics, potentially hindering its ability to generate the holistic, general-scene descriptions that this benchmark rewards.

In sharp contrast, Heron-bench demands a much deeper and more specialized understanding of Japanese culture. Its questions probe niche topics requiring fine-grained reasoning, such as knowledge of specific film directors, the context of historical documents, or details about less-famous regions. The decisive victory of the DEJIMA-Cap-All \& DEJIMA-VQA-All model here is a testament to our pipeline's effectiveness. We attribute this success to the synergistic combination of broad contextual cues from alt-text and precise, visually-grounded facts from object detection. This rich, multi-faceted data is essential for equipping the model with the detailed knowledge required to answer such expert-level questions, a task for which simpler descriptions are inadequate.

Ultimately, while strong performance on JA-VLM-Bench confirms robust general capabilities, the success on Heron-bench is particularly significant. It demonstrates that our All pipeline enables a model to progress beyond superficial recognition of cultural icons towards a genuinely deep and nuanced understanding of Japan. This capacity for grounded, in-depth reasoning on specialized cultural knowledge represents a more challenging and valuable achievement, showcasing the true potential of the DEJIMA dataset.
\begin{table}[t]
  \centering
  \resizebox{\linewidth}{!}{
    \begin{tabular}{llrr}
      \toprule
      Stage 1              & Stage 2                & \shortstack{JA-VLM-Bench-In-the-Wild                  \\(LLM-as-a-Judge $\uparrow$)} & \shortstack{Heron-bench\\(LLM-as-a-Judge $\uparrow$)} \\
      \midrule
      STAIR Captions       & Japanese Visual Genome & 3.04                                 & 31.57          \\
      STAIR Captions       & GQA Translation        & 1.58                                 & 20.54          \\
      MS COCO Translation  & Japanese Visual Genome & 2.88                                 & 33.94          \\
      DEJIMA-Cap-Simple    & DEJIMA-VQA-Refined     & \textbf{3.12}                        & 44.82          \\
      DEJIMA-Cap-Refined   & DEJIMA-VQA-Refined     & 1.96                                 & 15.89          \\
      DEJIMA-Cap-Detection & DEJIMA-VQA-Detection   & 1.36                                 & 21.95          \\
      DEJIMA-Cap-All       & DEJIMA-VQA-All         & 2.48                                 & \textbf{52.26} \\
      \bottomrule
    \end{tabular}
  }
  \caption{Evaluation on the learned model using Japanese V\&L benchmarks.}
  \label{tab:experimental_results_actual}
\end{table}

\subsection{Human Evaluation of VLM Outputs}
In addition to automatic evaluations, we conducted human evaluations of VLM outputs on Heron-bench to complement the benchmark results.
All outputs for 103 questions were manually assessed following the same protocol as the dataset-level evaluation described in Section~\ref{sec:dataset-analysis}.
Each output was evaluated by one Japanese-speaking crowd worker who compared two model outputs side by side (e.g., between models trained on different dataset pipelines), with randomized display order to avoid positional bias.
A total of 412 samples (103 questions × 4 model comparisons) were evaluated by 28 workers in total.
Each worker was assigned 15 samples, except for four workers who were assigned 13 samples, and one identical control item was inserted for quality checking.
Annotations from workers whose responses to the control were inconsistent were excluded, accounting for approximately 21.4\% of workers.
The resulting effective sample size \( n \) for each comparison is reported in Table~\ref{tab:heron-bench-heron-bench-combined}.

Before the evaluation, all workers were presented with the definitions of the evaluation criteria for pairwise model comparison.
The four metrics were as follows: \emph{Japaneseness of text}, which measures the presence of Japanese-specific expressions and cultural contexts; \emph{Naturalness of text}, which evaluates the fluency and naturalness of the Japanese language; \emph{Q–A relevance}, which measures whether the answer appropriately addresses the question’s intent; and \emph{Answer correctness}, which assesses factual accuracy given the image and question.

As shown in Table~\ref{tab:heron-bench-heron-bench-combined},
the model trained on the \textbf{DEJIMA-Cap-All \& DEJIMA-VQA-All} pipeline achieved the highest preference in nearly all metrics,
including \emph{Japaneseness of text}, Naturalness, and notably \emph{Answer correctness}.
This indicates that the linguistic and cultural richness of DEJIMA datasets translates directly into improved downstream VLM performance.

When compared to the MS COCO Translation \& Japanese Visual Genome pipeline,
the All-pipeline outputs were preferred overwhelmingly (97.5\% for Japaneseness, 92.5\% for Naturalness), confirming the strong impact of using Japanese-grounded data.
Against ablated versions, alt-only or detection-only pipelines showed lower scores, while the combined pipeline consistently achieved the best balance of fluency and factuality.
Notably, the Detection-based variant performed well in \emph{Q–A relevance} and \emph{Answer correctness}, while the Refined variant excelled in Naturalness, again supporting the complementary relationship between the two information sources.

\begin{table}[t]
  \centering
  \resizebox{\linewidth}{!}{
    \begin{tabular}{llllll}
      \toprule
      Compared Pipeline                              & $n$ & \shortstack{Japaneseness                                                                                  \\of text}     & \shortstack{Naturalness\\of text}      & \shortstack{Q-A                                     \\relevance}            & \shortstack{Answer\\correctness}       \\
      \midrule
      MS COCO Translation \&  Japanese Visual Genome & 75  & 97.33\textsuperscript{*} & 92.00\textsuperscript{*} & 61.33                    & 77.33\textsuperscript{*} \\
      DEJIMA-Cap-Simple \&    DEJIMA-VQA-Refined     & 88  & 79.55\textsuperscript{*} & 79.55\textsuperscript{*} & 53.41                    & 50.00                    \\
      DEJIMA-Cap-Detection \& DEJIMA-VQA-Detection   & 73  & 89.04\textsuperscript{*} & 87.67\textsuperscript{*} & 84.93\textsuperscript{*} & 90.41\textsuperscript{*} \\
      DEJIMA-Cap-Refined \&   DEJIMA-VQA-Refined     & 88  & 63.64\textsuperscript{*} & 97.73\textsuperscript{*} & 77.27\textsuperscript{*} & 70.45\textsuperscript{*} \\
      \bottomrule
    \end{tabular}
  }
  \caption{VLM output: pairwise preference vs. \textbf{DEJIMA-Cap-All \& DEJIMA-VQA-All} pipeline. * indicates significance at 5\%.}
  \label{tab:heron-bench-heron-bench-combined}
\end{table}

\subsection{VLM Output Example}

\begin{figure}[t]
  \centering
  \includegraphics[width=0.9\textwidth]{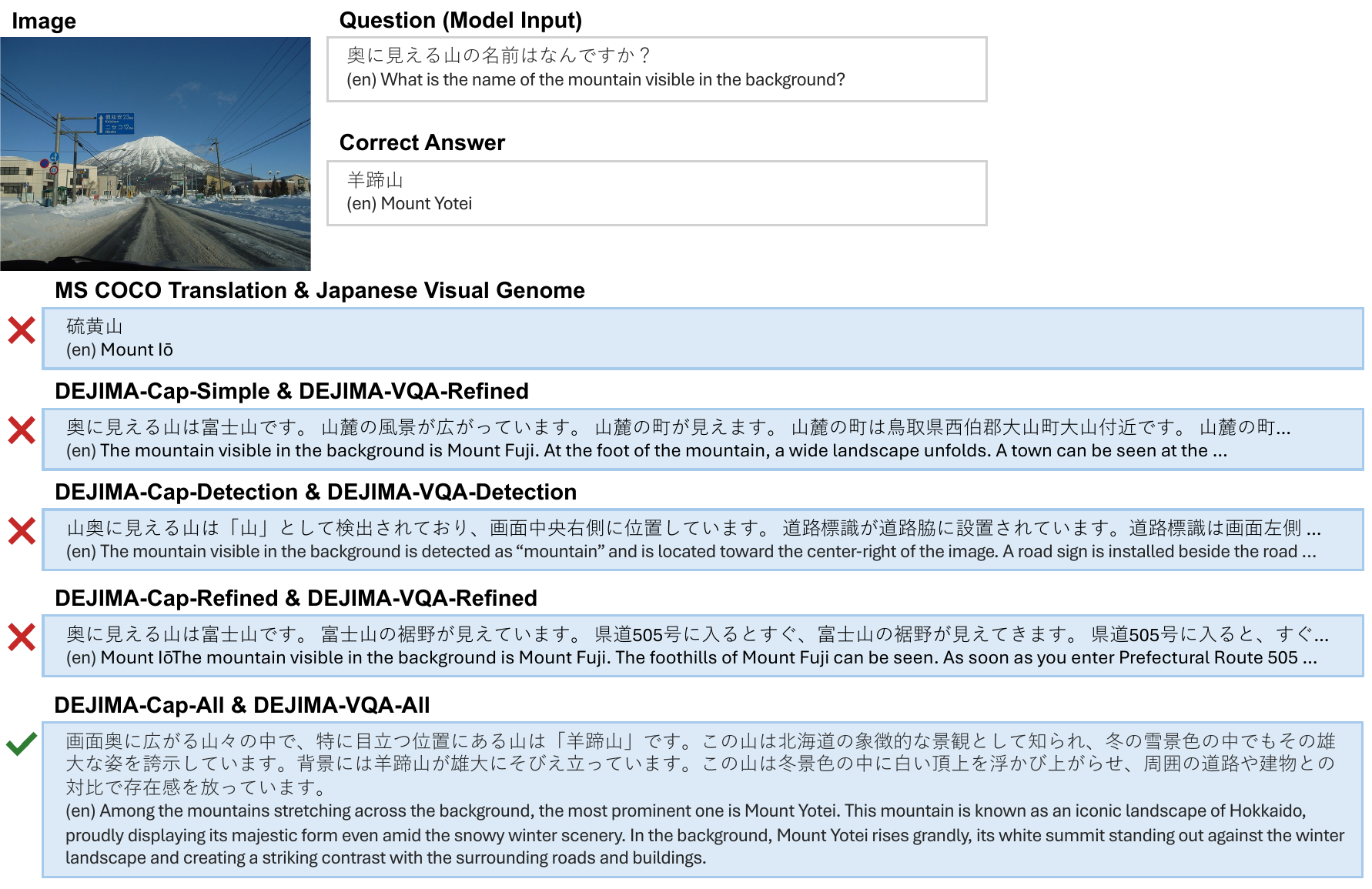}
  \caption{VLM Output Example for the question ``What is the name of the mountain visible in the background?''
    We compare outputs from five pipelines against the correct answer (Mount Yotei).
    The All pipeline (alt-text + detection) yields correct identification with natural phrasing and explicit visual grounding,
    while translation-based and ablated variants either hallucinate specific locations or stay generic despite object detections.
    Extremely long or repetitive model outputs have been truncated for clarity.}
  \label{fig:vlm-output}
\end{figure}

Figure~\ref{fig:vlm-output} presents a qualitative example comparing generations across training pipelines for the question
``What is the name of the mountain visible in the background?''
The correct answer is ``Mount Yotei.'' We show outputs from five representative pipelines.
As illustrated, baseline and ablated variants tend to either misidentify the mountain (e.g., producing a popular landmark) or remain generic despite object detections.
In contrast, the All pipeline correctly names \textit{Mount Yotei} and provides a visually grounded rationale that refers to salient scene attributes.

\section{Conclusion}
We introduced DEJIMA, a detection-guided and LLM-refined pipeline for constructing culturally grounded Japanese Vision-and-Language resources at scale.
Our approach yields two datasets—DEJIMA-Cap and DEJIMA-VQA—each with $\sim$3.88M entries, surpassing existing Japanese resources by more than an order of magnitude while preserving linguistic naturalness and visual grounding.
Human evaluations showed consistent gains in \emph{Japaneseness} and \emph{naturalness} over translation- and manual-annotation baselines, and downstream experiments demonstrated strong improvements on Japanese multimodal benchmarks (JA-VLM-Bench-In-the-Wild and Heron-bench).

\section{Ethics Statement}

This work involves the large-scale collection and automatic refinement of Japanese image–text data from the public web and the use of paid human evaluations. We describe the ethical considerations across data sourcing, licensing and redistribution, privacy, annotator welfare, cultural sensitivity, safety, transparency, and intended use.

\paragraph{Data sourcing and licensing.}
We collect candidate image–alt-text pairs from Common Crawl and retain only Japanese pages after language and adult filtering. Web content may be copyrighted or subject to site-specific terms. To respect rights of content owners, we do not redistribute images; we release only non-image artifacts (e.g., URLs, text, derived metadata, and model-generated text). Users of the resource must download images directly from origin servers and abide by the original licenses and terms of service. Upon publication, we will provide a takedown channel; we will promptly remove or update entries upon substantiated requests from rights holders.

\paragraph{Privacy and personal data.}
Public web images can contain people or private locations. Our pipeline does not attempt to identify individuals, infer sensitive attributes, or link content to specific persons. We apply content filters to reduce exposure to adult content and other unsafe material and exclude pages that are likely to contain personally identifying information in the text layer. We do not release any additional annotations about identities or sensitive attributes. We ask downstream users to avoid tasks such as face recognition, re-identification, or surveillance, and to comply with applicable privacy laws in their jurisdictions.

\paragraph{Safety filtering and hallucination control.}
To mitigate propagation of harmful or unverifiable claims during LLM refinement, generation is conditioned on detected visual evidence and guided by constraints that discourage unsupported statements.
In addition, we performed text-based safety filtering at the alt-text stage to remove entries containing adult, violent, or otherwise harmful content before downstream processing.

\paragraph{Cultural sensitivity and bias.}
The dataset emphasizes Japanese content. While this focus is central to our goal of culturally grounded Japanese V\&L resources, it can also amplify culture-specific biases or stereotypes. We explicitly caution against using the resource to make normative claims about Japanese culture and encourage researchers to conduct bias and fairness analyses for their particular use cases.

\paragraph{Annotator welfare.}
Human evaluations were conducted with consenting, Japanese-speaking crowd workers. Instructions detailed the task, quality expectations, and the right to withdraw. We compensated workers at or above local minimum-wage equivalents for the expected time per assignment, paid for approved tasks, and minimized exposure to potentially sensitive content through pre-filtering. No personally identifying information about workers is collected or released.

\paragraph{Environmental considerations.}
Our pipeline is designed to be compute-efficient by leveraging detection-guided prompts and batching. We report approximate compute used for dataset construction and model training to support assessments of environmental impact and reproducibility. We encourage downstream users to consider energy-efficient settings and to reuse our released artifacts to avoid redundant computation.

\paragraph{Intended use and misuse.}
The resource is intended for research and development of multimodal systems that generate or understand Japanese text and imagery. It should not be used for surveillance, identity inference, discriminatory decision-making, medical or legal advice, or any context where errors could cause harm without appropriate human oversight. We release documentation (datasheet/model card), licensing guidance, and known limitations to support responsible deployment.

\paragraph{Compliance.}
This study uses publicly available web data and non-identifiable human annotations. No institutional review board approval was required under our local guidelines; we nonetheless adhered to principles of informed consent, data minimization, and respect for rights holders and annotators. Downstream users are responsible for ensuring compliance with local laws and the original content licenses when using the released artifacts.

\section{Limitations}

While DEJIMA presents a scalable and culturally grounded approach to constructing large-scale Japanese Vision-and-Language datasets, several limitations remain.

First, although we applied extensive filtering and grounding mechanisms, web-derived alt-text may still contain noise, inaccuracies, or cultural bias. Image–text consistency, in particular, is not always guaranteed and can be weaker than human annotation in some cases. Detection-guided grounding reduces unverifiable or hallucinated statements, but challenges persist for long-tail cultural references and abstract descriptions.

Second, our concept of “Japaneseness” was intentionally left open-ended and was evaluated based on annotators’ subjective judgment rather than a predefined operational definition. A more systematic or multidimensional framework for measuring cultural grounding would enhance reproducibility and interpretability.

Third, the current work focuses exclusively on Japanese data. However, the proposed pipeline is language-agnostic and could, in principle, be applied to other languages by replacing the language filters and refinement models. Extending the approach to multilingual settings remains an important direction for future work.

Finally, while we constructed datasets of approximately 3.88 million pairs, the pipeline is inherently scalable. Running additional iterations with expanded web sources could further enlarge the dataset. Future work should also investigate efficient quality control at even larger scales and assess environmental and computational impacts.

\section{Data and Code Availability}

To promote transparency and reproducibility, we will release the full dataset construction pipeline, metadata, and evaluation scripts upon publication.

Specifically, we will provide:
(1) source code for the data collection, filtering, deduplication, detection, and LLM-based refinement stages;
(2) JSON-formatted metadata containing image URLs, alt-texts, alignment scores, detection outputs (labels and bounding boxes), and LLM-generated texts;
and (3) configuration files and prompts used for generation and evaluation.

Following the licensing terms of the original web data, we will not redistribute image files themselves.
Instead, we release URL lists and derived text annotations so that users can reproduce the dataset by downloading images directly from their original sources, respecting each website’s terms of service.

All released components (code, metadata, and annotations) will be available under a permissive open license allowing both research and commercial use, except where restricted by upstream content licenses.
We will also provide detailed documentation (datasheet) describing the data sources, filtering criteria, model configurations, and ethical considerations to ensure responsible use and facilitate future extensions.

The resources will be released after the review period to preserve anonymity.

\section*{Acknowledgements}
This work was partially supported by JST Moonshot R\&D Grant Number JPMJPS2011, CREST Grant Number JPMJCR2015 and Basic Research Grant (Super AI) of Institute for AI and Beyond of the University of Tokyo.

\bibliographystyle{unsrt}
\bibliography{references}  

\end{document}